\documentclass[sigconf]{acmart}

\usepackage{threeparttable}
\usepackage[utf8]{inputenc} % allow utf-8 input

\usepackage{booktabs}       % professional-quality tables
\usepackage{nicefrac}       % compact symbols for 1/2, etc.
\usepackage{microtype}      % microtypography
\usepackage{xcolor}         % colors
\usepackage{soul}
\usepackage{url}
\usepackage[font={small,bf}]{caption}
\usepackage{graphicx}
\usepackage{amsmath}
\usepackage{amsthm}
\usepackage{algorithm}
\usepackage{algorithmic}
\usepackage{multirow}
\usepackage{subfig}
\usepackage{bm}
\usepackage{color}

\usepackage{tablefootnote}
\usepackage{lipsum}
\usepackage{subcaption}
\usepackage{subfig}
\usepackage{float} 
\usepackage{tcolorbox}

\usepackage{xspace}
\usepackage{stfloats}
\usepackage{marvosym}
\usepackage{enumitem}
\usepackage{pifont}%http://ctan.org/pkg/pifont

\usepackage{color, colortbl}
\AtBeginDocument{%
  }

\copyrightyear{2026}
\acmYear{2026}
\setcopyright{cc}
\setcctype{by}
\acmConference[SIGIR '26]{Proceedings of the 49th International ACM SIGIR Conference on Research and Development in Information Retrieval}{July 20--24, 2026}{Melbourne, VIC, Australia}
\acmBooktitle{Proceedings of the 49th International ACM SIGIR Conference on Research and Development in Information Retrieval (SIGIR '26), July 20--24, 2026, Melbourne, VIC, Australia}
\acmDOI{10.1145/3805712.3809983}
\acmISBN{979-8-4007-2599-9/2026/07}
\acmSubmissionID{sp092}

\begin{document}
% \settopmatter{printacmref=false} % Removes citation information below abstract
% \renewcommand\footnotetextcopyrightpermission[1]{} % removes footnote with conference information in first column
% \pagestyle{plain} % removes running headers
\title{Understanding DNNs in Feature Interaction Models: A Dimensional Collapse Perspective}

% \affiliation{%

\author{Jiancheng Wang}
\orcid{0009-0005-3532-3699}
\email{wangjc830@mail.ustc.edu.cn}
\affiliation{%
  \institution{University of Science and Technology
of China \& State Key Laboratory of
Cognitive Intelligence}
  \state{Hefei}
  \country{China}}

\author{Mingjia Yin}
\orcid{0009-0005-0853-1089}
\email{mingjia-yin@mail.ustc.edu.cn}
\affiliation{%
  \institution{University of Science and Technology
of China \& State Key Laboratory of
Cognitive Intelligence}
  \state{Hefei}
  \country{China}}

\author{Hao Wang}
\authornote{Corresponding author.}
\orcid{0000-0001-9921-2078}
\email{wanghao3@ustc.edu.cn}
\affiliation{%
  \institution{University of Science and Technology
of China \& State Key Laboratory of
Cognitive Intelligence}
  \state{Hefei}
  \country{China}}

\author{Enhong Chen}
\orcid{0000-0002-4835-4102}
\email{cheneh@ustc.edu.cn}
\affiliation{%
  \institution{University of Science and Technology
of China \& State Key Laboratory of
Cognitive Intelligence}
  \state{Hefei}
  \country{China}}



% \setlength{\parindent}{0pt}

\begin{abstract}
DNNs have gained widespread adoption in feature interaction recommendation models.
However, there has been a longstanding debate on their roles.
On one hand, some works claim that DNNs possess the ability to implicitly capture high-order feature interactions. 
Conversely, recent studies have highlighted the limitations of DNNs in effectively learning dot products, specifically second-order interactions, let alone higher-order interactions.
In this paper, we present a novel perspective to understand the effectiveness of DNNs: \emph{their impact on the dimensional robustness of the representations}. 
In particular, we conduct extensive experiments involving both parallel DNNs and stacked DNNs. 
Our evaluation encompasses an overall study of complete DNN on two feature interaction models, alongside a fine-grained ablation analysis of components within DNNs. 
Experimental results demonstrate that \emph{both parallel and stacked DNNs can effectively mitigate the dimensional collapse of embeddings.} 
Furthermore, a gradient-based theoretical analysis, supported by empirical evidence, uncovers the underlying mechanisms of dimensional collapse.
The code is accessible for reproduction.\footnote{\url{https://github.com/USTC-StarTeam/Dimensional-Collapse-Analysis}}
\end{abstract}





\maketitle

\section{Introduction}

% The online advertising industry has experienced tremendous growth, evolving into a billion-dollar business. It is projected to generate an annual revenue of 225 billion US dollars between 2022 and 2023, with a year-on-year increase of 7.3\% ~\cite{iab-report}. 
% In this industry, a key challenge revolves around delivering targeted advertisements to the appropriate customers within a given context.
% Accurately predicting the click-through rate (CTR) plays a vital role in addressing this challenge and has garnered significant attention over the past decade ~\cite{FM2010, FFM2016, FwFM2018, xDeepFM2018, DLRM2019, FmFM2021, DCNv22021, zhang2022dhen, FinalMLP2023, zhu2023final,pan2024ads, wang2025enhancing, yinfeature}.

Click-through rate (CTR) prediction plays a core role in online advertising and recommendation systems~\cite{DLRSSurvey2019, ctr_survey_2021, pan2024ads}. 
CTR prediction tasks commonly involve the handling of multi-field categorical data~\cite{zhang2016deep, FwFM2018} to achieve accurate predictions. 
The primary challenge lies in effectively capturing interactions between these features, which is complicated by the extreme sparsity of feature co-occurrence due to the high cardinality of features and the rarity of user feedback.
To tackle this challenge, various approaches have been proposed, aiming to explicitly model feature interactions~\cite{FM2010, FibiNet2019, FmFM2021, zhang2024wukong} or hybrid methods that combine explicit modeling with deep neural networks (DNN)~\cite{PNN2016, xDeepFM2018, AutoInt2019, DCNv22021, FinalMLP2023}. 
By leveraging these interaction modules, the models can effectively capture and learn the complex relationships and interactions between features, enabling more accurate predictions.

\begin{figure}[t]
    \centering
    \subfloat[Parallel DNN]{%
        \includegraphics[width = 0.24\textwidth]{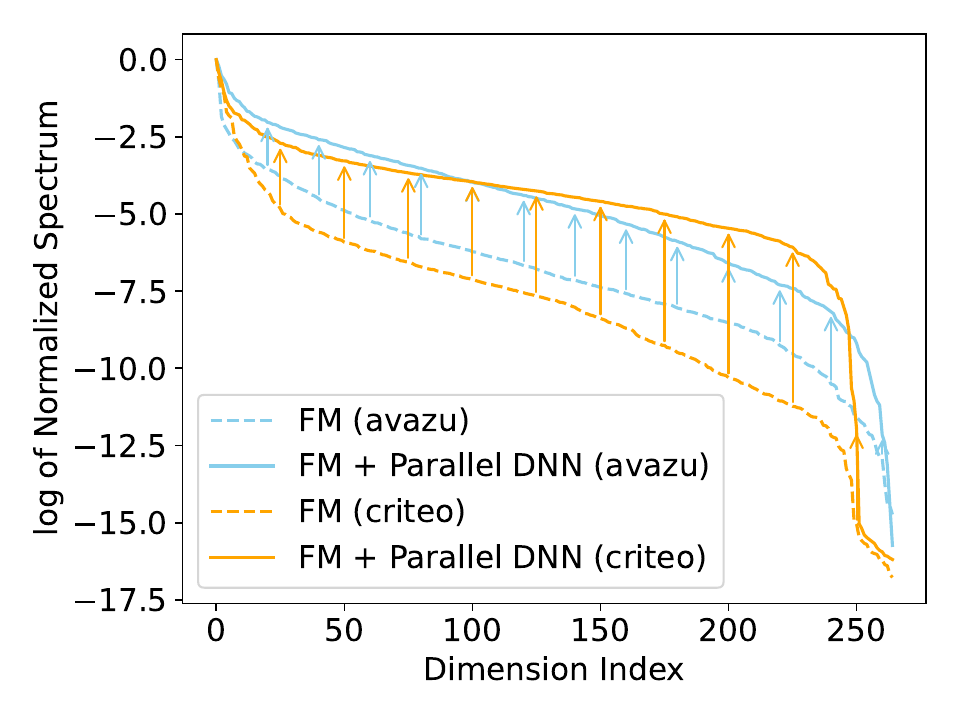}
        \label{fig:intro_para}
        }
    \subfloat[Stacked DNN]{
        \includegraphics[width = 0.24\textwidth]
        {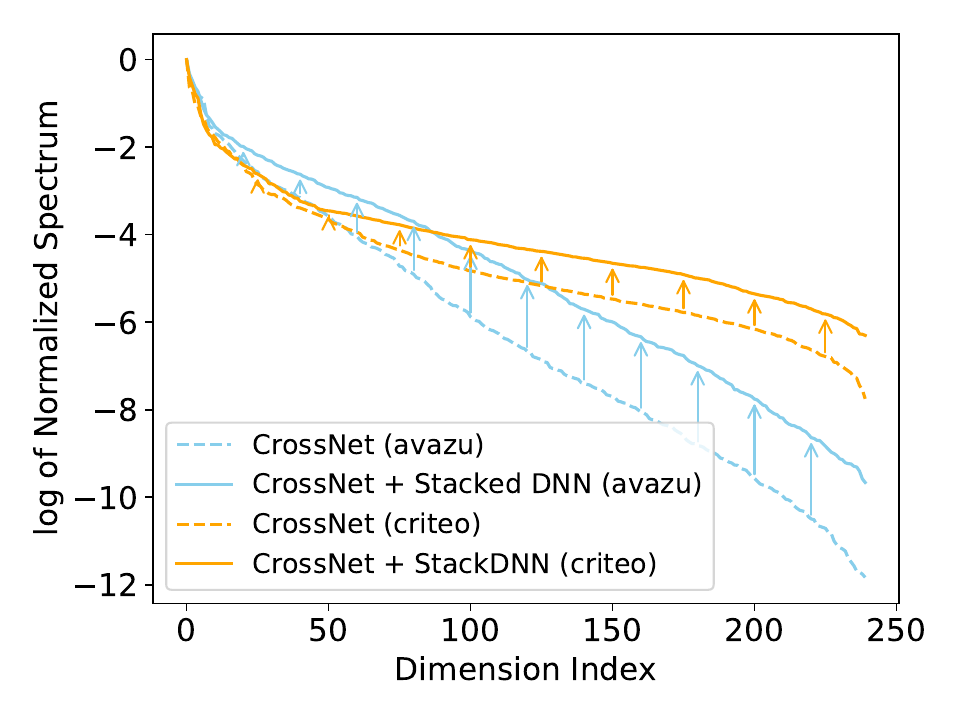}
        \label{fig:intro_stack}
    }
\caption{We investigate the role of DNN in feature interaction models from a novel embedding dimensional collapse mitigation perspective: both parallel DNN and stacked DNN can improve the singular value spectrum (y-axis), hence alleviating dimensional collapse of models FM and CrossNet on two CTR datasets.}
\vspace{-15pt}
\label{fig:intro}
\end{figure}

% \begin{figure}[t]
%     \centering
%     \begin{subfigure}[b]{0.24\textwidth} 
%         \includegraphics[width=\textwidth]{figure/Structure/spectrum_intro_para.pdf}
%         \caption{Parallel DNN}
%         \label{fig:intro_para}
%     \end{subfigure}
%     % \hfill
%     \begin{subfigure}[b]{0.24\textwidth} 
%         \includegraphics[width=\textwidth]{figure/Structure/spectrum_intro_stack.pdf}
%         \caption{Stacked DNN}
%         \label{fig:intro_stack}
%     \end{subfigure}
%     \caption{We investigate the role of DNN in feature interaction models from a novel embedding dimensional collapse mitigation perspective: both parallel DNN and stacked DNN can improve the singular spectrum (y-axis), hence alleviating embeddings dimensional collapse of feature interaction models such as FM and CrossNet on two CTR datasets.}
%     \label{fig:intro}
% \end{figure}

Numerous studies have incorporated deep neural networks (DNN) in CTR prediction tasks, typically by applying them in parallel with explicit feature interaction modules ~\cite{WideAndDeep2016, DeepFM2017, DCNv22021, xDeepFM2018} or by stacking them on top of outputs ~\cite{NFM2017, DCNv22021, FinalMLP2023}. 
These works generally claim that DNN contributes by capturing high-order implicit feature interactions.
However, this assumption has been the subject of ongoing debate.
For instance, prior studies ~\cite{beutel2018latent, NCF_Revisited2020} highlight the intrinsic limitations that learning dot product with a multi-layer perceptron (MLP) can be non-trivial. 
Furthermore, DCNv2~\cite{DCNv22021} showed that a ReLU-based DNN can underperform high-order explicit interaction models such as CrossNet. 
Despite the empirical success of DNNs, the underlying reasons driving their effectiveness in recommendation models remain insufficiently understood.

% Meanwhile, in representation learning~\cite{UnderstandingDimensionCollapse2021}, \textbf{dimensional collapse} has been increasingly recognized as a critical bottleneck that limits the expressiveness of embeddings, where embeddings degenerate into a narrow, low-dimensional subspace of the representation space.
% This phenomenon is characterized by rapid singular value decay in the embedding matrix, as shown in Fig.\ref{fig:intro}.
% Follow-up studies ~\cite{qiu2022contrastive, zhang2023mitigating, chen2024towards, he2024preventing} have proposed various strategies(e.g. regularization~\cite{bardes2021vicreg}) to assess and alleviate collapse，这些工作提升我们理解dimensional collapse对表征的影响.
Meanwhile, in representation learning~\cite{UnderstandingDimensionCollapse2021}, \textbf{dimensional collapse} has been increasingly recognized as a critical bottleneck that limits the expressiveness of embeddings, where embeddings degenerate into a narrow, low-dimensional subspace of the representation space.
This phenomenon can be characterized by the rapid decay of singular values in the embedding matrix, as shown in Fig.\ref{fig:intro}.
Follow-up studies have proposed various strategies (\emph{e.g.} regularization~\cite{bardes2021vicreg}) to diagnose and mitigate collapse~\cite{qiu2022contrastive, zhang2023mitigating, chen2024towards, he2024preventing}, advancing our understanding of how collapse affects representation quality.
% and proposed effective techniques to tackle it.

In recommender systems, dimensional collapse is also a significant restriction to achieve robust representations. 
Inspired by prior studies in contrastive learning~\cite{wu2021self, mao2021simplex, UnderstandingDimensionCollapse2021, yu2022graph}, which demonstrate that less-collapsed embeddings yield superior model performance, recent efforts have begun to address the impact of dimensional collapse from various aspects, including objective functions~\cite{zhang2023mitigating, chen2024towards} and architectural designs~\cite{FinalMLP2023, guo2023multi-embedding, pan2024ads, yin2025GE4Rec}. 
Collectively, these works highlight that addressing dimensional collapse provides a crucial perspective for enhancing the robustness of representations in recommender systems.

% In recommender systems, previous work~\cite{guo2023multi-embedding} formalized this phenomenon through the proposed \emph{Interaction-Collapse Theory}, which reveals that interacting with low-information-abundance fields tends to induce collapsed embeddings.

Motivated by these, we revisit the role of DNNs in recommendation models, particularly within feature interaction modules, from the perspective of dimensional collapse. 
We present a comprehensive understanding of DNNs through theoretical and empirical analysis.
Through systematic experiments on both parallel and stacked DNNs, we demonstrate that DNNs can effectively mitigate dimensional collapse.
Based on experimental results, we conduct a theoretical and empirical analysis of gradients, revealing the underlying mechanisms by which DNNs alleviate this issue.
In summary, our contributions can be summarized as follows:
\begin{itemize}[leftmargin=*] 
    \item To the best of our knowledge, we are the first to explore the role of DNNs in feature interaction models from the dimensional collapse perspective.
    \item We find that both parallel and stacked DNNs alleviate dimensional collapse in pure feature interaction model FM and CrossNet on two public datasets, shown in Fig.\ref{fig:intro}, revealing a new effect of DNNs beyond implicit feature interaction.
    \item We investigate the effects of the two components in the DNN and observe that both the linear DNN component and the non-linear activations contribute to alleviating dimensional collapse. 
    \item We provide a gradient-based analysis to illustrate how DNNs mitigate dimensional collapse.
\end{itemize}

% \begin{figure*}[t]
%     \centering
%     \includegraphics[width=0.8\linewidth]{figure/main.pdf}
%     \caption{Illustration of DNN integration in feature interaction models: (a) depicts parallel DNN structure, (b) represents stacked DNN structure. Progressive incorporating linear DNN and non-linear activation components gradually mitigates the dimensional collapse of embeddings.}
% % \caption{Model architecture illustration of plain feature interaction model, \emph{i.e.} (a), employing both parallel and stacked non-linear DNN upon the feature interaction model, \emph{i.e.} (c) and (e), as well as variants with only linear DNN, \emph{i.e.} (b) and (d).\textcolor{red}{Todo: rewrite}}
% \label{fig:overall_structure}
% \end{figure*}

\section{Background}

% \subsection{Feature Interaction Models}
% In this paper, we investigate two representative feature interaction models: FM (Factorization Machines)~\cite{FM2010} as a classic second-order interaction model and CrossNet (Cross Network within DCN V2)~\cite{DCNv22021} as a classic high-order interaction model.

% \paragraph{FM.} FM calculates the interactions between all feature embeddings using inner products. Formally, it can be expressed as:
% \begin{equation}
%     \Phi_{FM} = \sum_i \sum_j x_i x_j \langle \mathbf{v}_i, \mathbf{v}_j \rangle,
% \end{equation}
% here $x_i$ and $x_j$ represent the values of feature $i$ and $j$, respectively, and $\mathbf{v}_i$ and $\mathbf{v}_j$ denote the corresponding feature embeddings. $\langle \cdot\rangle$ represents dot product.

% \paragraph{CrossNet.} CrossNet explicitly models high-order feature interactions by recursively combining the input embedding with itself. Formally, the $l$-th layer output is computed as:
% \begin{equation}
%     \mathbf{h}_l = \mathbf{h}_0 \odot (W_{l-1} \mathbf{h}_{l-1} + b_{l-1}) + \mathbf{h}_l,
% \end{equation}
% where $\mathbf{h}_0$ is the concatenated input embedding, denoted as $[\dots,\mathbf{v}_i, \dots]$, $\odot$ denotes element-wise multiplication and $W_{l-1}$ and $b_{l-1}$ denotes the field-pair wise projection matrix and the bias vector, respectively, for the $l-1$-th layer.

\subsection{Problem Definition}

Click-Through Rates (CTR) prediction aims to predict the probability of a user clicking on an advertisement in advertising systems.
A typical CTR prediction problem can be formally defined as $\{ \mathcal{X}, \mathcal{Y} \}$, where $\mathcal{X}$ is a multi-categorical feature set, while $\mathcal{Y} \in \{0, 1\}$ is a label set indicating whether users have clicked the advertisement.
Usually, $\mathcal{X}$ includes user, item, and contextual features.
Then a recommendation model aims to learn a mapping from $\mathcal{X}$ to $\mathcal{Y}$.

\subsection{Dimensional Collapse Evaluation Protocols} \label{sec:evaluation}
To evaluate the degree of dimensional collapse in embeddings, we utilize the singular value spectrum to observe the distribution macroscopically and employ the RankMe metric for quantitative evaluation.

\noindent \textbf{Singular Value Spectrum. }
To diagnose dimensional collapse of feature embedding, previous work suggest analyzing the singular value spectrum of representation~\cite{UnderstandingDimensionCollapse2021, zhang2023mitigating, he2024preventing, yin2025GE4Rec}.
The embedding matrix $\mathbf{Z}\in \mathbb{R}^{B\times D}$ is collected on the test set, where $B$ denotes the batch size and $D$ denotes the embedding dimension. 
By performing Singular Value Decomposition (SVD) on the covariance matrix $\mathbf{C}\in \mathbb{R}^{D\times D}$ of $\mathbf{Z}$, derived as $\mathbf{C}=\frac{1}{B} \sum_{i=1}^B\left(\mathbf{z}_i-\overline{\mathbf{z}}\right)\left(\mathbf{z}_i-\overline{\mathbf{z}}\right)^T$, with $\overline{\mathbf{z}}=\frac{1}{B} \sum_{i=1}^B \mathbf{z}_i$.
We obtain the singular values $\boldsymbol{\sigma} = \{\sigma_1, \sigma_2, \dots, \sigma_d\}$ sorted in descending order.
From Fig. \ref{fig:intro}, a rapidly decaying spectrum indicates that most of the embedding information is concentrated in a small number of dimensions, implying that the representation effectively collapses into a low-dimensional subspace.
% The singular value spectrum of the embedding space is used in ~\cite{UnderstandingDimensionCollapse2021} to evaluate the phenomenon of dimensional collapse.
% We measure the dimensionality by collecting the embedding matrix $\mathbf{Z}\in \mathbb{R}^{B\times D}$ on the validation dataset, where $B$ denotes the batch size and $K$ denotes the embedding dimension. 
% We compute the covariance matrix $\mathbf{C}\in \mathbb{R}^{D\times D}$ of the embeddings, derived as $\mathbf{C}=\frac{1}{B} \sum_{i=1}^B\left(\mathbf{z}_i-\overline{\mathbf{z}}\right)\left(\mathbf{z}_i-\overline{\mathbf{z}}\right)^T$, with $\overline{\mathbf{z}}=\frac{1}{B} \sum_{i=1}^B \mathbf{z}_i$.
% Eventually, we visualize the distribution of singular values $S={diag}\left(\sigma^k\right)$ via singular value decomposition(SVD) ($\mathbf{C}=U S V^T$), normalize them by the maximum singular value and logarithmic scale $\left(\left\{\log \left(\frac{\sigma^k}{\sigma_{max}}\right)\right\}\right)$. 
% From Fig. \ref{fig:intro}, we observe that a number of singular values collapse to nearly zero, thus representing collapsed dimensions.

\noindent \textbf{RankMe. }
To quantify the robustness of representations, we employ RankMe \cite{garrido2023rankme}, a smooth estimator of matrix rank defined as the Shannon entropy~\cite{shannon1951entropy} of the normalized singular values of the representation matrix $\mathbf{Z} \in \mathbb{R}^{N \times K}$. It effectively assesses the uniformity of the singular value distribution.
\begin{equation}
    RankMe(Z) = exp(-\sum_{k=1}^{\min(N, K)}p_k \log p_k), \quad p_k = \frac{\sigma_k(Z)}{\|\sigma(Z)\|_1},
\end{equation}
where $\sigma_k(Z)$ represents the $k$-th singular value of $Z$, and $\|\cdot\|_1$ denotes the $L_1$ norm. 
A higher RankMe value indicates more robust and less-collapsed representations.

Building on these evaluation protocols, we investigate the role of DNNs in feature interaction models from the perspective of dimensional collapse.

\subsection{Setup}
\noindent \textbf{Datasets \& Baselines. }
Experiments are conducted on two widely used public CTR datasets: Avazu~\cite{avazu} and Criteo~\cite{criteo}.
For baselines, we consider two representative pure feature interaction backbones:
second-order explicit interaction model \textbf{FM}~\cite{FM2010}, along with its DNN-augmented variants DeepFM~\cite{DeepFM2017} and NFM~\cite{NFM2017}; high-order interaction model \textbf{CrossNet}, together with its DNN-enhanced variants DCNv2~\cite{DCNv22021}.

\noindent \textbf{Metric. }  We adopt \textbf{AUC} to evaluate predictive performance and employ \textbf{RankMe} to quantify the degree of dimensional collapse in feature embeddings.
For spectral analysis, we collect 50000 random samples in the validation dataset.

% \paragraph{\textbf{Implementation}}
\noindent \textbf{Implementation. }
To ensure a fair comparison, we strictly follow the optimal configurations recommended by FuxiCTR~\cite{FuxiCTR2020, BARS2022}, a widely adopted open-source CTR benchmark library. 
The embedding dimension $d$ is set to 16 for Avazu and 10 for Criteo. 
For the DNN components, on the Avazu dataset, the parallel DNN uses a 4-layer MLP with 2000 hidden units, while the stacked DNN employs a 3-layer MLP with 400 units.

\begin{figure*}[t]
    \centering
    \includegraphics[width=1.0\linewidth]{figure/structure.png}
    \caption{Illustration of DNN integration in feature interaction models: (a) depicts parallel DNN structure, (b) represents stacked DNN structure. Progressive incorporating linear DNN and non-linear activation components gradually mitigates the dimensional collapse of embeddings.}
\label{fig:overall_structure}
\end{figure*}

\section{Revisited Effectiveness of DNNs from the Dimensional Collapse Perspective}
In this section, we revisit the role of DNNs in feature interaction models from the perspective of dimensional collapse. We aim to understand how DNNs reshape the embedding space and alleviate dimensional collapse. 
Specifically, as shown in Fig.~\ref{fig:overall_structure}, the integration of DNNs into feature interaction models can be broadly categorized into two architectures:
(a) \textbf{Parallel DNN (p-DNN)}, where the DNN receives the same input as the interaction module and operates in parallel;
(b) \textbf{Stacked DNN (s-DNN)}, where the DNN is placed on top of the interaction module and takes its output as input.

For clarity and systematic analysis, we investigate these two integration strategies separately. Specifically, we conduct experiments to investigate the following research questions:
\begin{itemize}[leftmargin=*]
    \item RQ1: How does DNN as a whole affect both predictive performance and the dimensional robustness of feature interaction models FM and CrossNet?
    \item RQ2: DNN can be decomposed into linear DNNs and non-linear activations. What are the respective contributions of these two components in mitigating dimensional collapse?
\end{itemize}

\subsection{RQ1: Overall Impact of Whole DNN}
We examine the impact of integrating parallel and stacked DNN into feature interaction model FM and CrossNet. This covers well-known architectures such as DeepFM (FM+p-DNN), NFM (FM+s-DNN), parallel DCNv2 (CrossNet+p-DNN), and stacked DCNv2 (CrossNet+s-DNN).
The study is conducted from two aspects: predictive performance and dimensional robustness of embeddings concatenated from all feature fields.

\noindent \textbf{Predictive Performance. }
As shown in Tab. \ref{tab:dnn}, adding a DNN component consistently improves AUC across both datasets. While p-DNN yields higher AUC on FM, s-DNN shows superior performance on CrossNet. These results align with prior benchmarks \cite{DeepFM2017, DCNv22021}, confirming that DNNs are essential for predictive accuracy.

\noindent \textbf{Dimensional Robustness. }
As quantified by RankMe in Tab. \ref{tab:dnn}, the improvements are substantial. For FM on Avazu, adding p-DNN and s-DNN boosts the RankMe score from 89.12 to 166.78 and 160.57, respectively. On the Criteo dataset, p-DNN increases FM's RankMe by 158\%, effectively rescuing the model from a low-rank state.

Furthermore, we visualize the spectrum of the logarithm of the normalized singular values, sorted in descending order.
As shown in Fig.~\ref{fig:intro}, we can observe that pure interaction models (FM/CrossNet) exhibit a sharp spectral decay, suggesting severe dimensional collapse. 
In contrast, integrating either p-DNN or s-DNN significantly flattens the spectrum, yielding consistently higher normalized singular values within the top 260 indices compared to the models without DNNs.
Specifically, as illustrated in Fig.~\ref{fig:intro_para}, p-DNN produces a more uniform singular value distribution, with normalized values increasing by an average of $10^5$ on the Avazu dataset and $10^3$ on Criteo. 
A similar trend is observed when applying s-DNN to CrossNet. 
We conclude the observations with the following result:

\begin{tcolorbox}[colback=blue!2!white,leftrule=2.5mm,size=title] 
    \emph{\textbf{Result 1}. Both parallel and stacked DNNs significantly mitigate dimensional collapse in pure feature interaction models, as evidenced by more uniform singular value spectra and substantially higher RankMe.} 
\end{tcolorbox}

\begin{table}[t]
    \centering
    \caption{Performance of integrating parallel DNN (p-DNN) / stacked DNN (s-DNN) on feature interaction model FM and CrossNet.}
    \begin{tabular}{c|c|cc|cc}
        \toprule
       \multirow{2}{*}{\textbf{Model}}  & \multirow{2}{*}{\textbf{Variants}}
        & \multicolumn{2}{c|}{\textbf{Avazu}} 
        & \multicolumn{2}{c}{\textbf{Criteo}} \\
        & 
        & AUC$\uparrow$ & RankMe$\uparrow$ 
        & AUC$\uparrow$ & RankMe$\uparrow$ \\
        \midrule
        \multirow{3}{*}{FM}
        & Base & 0.7848 & 89.12 & 0.8025 & 70.13 \\
        \cmidrule{2-6}
        & +p-DNN & \textbf{0.7927} & \textbf{166.78} & \textbf{0.8137} & \textbf{181.56} \\
        & +s-DNN & 0.7893 & 160.57 & 0.8050 & 72.93 \\
        \midrule
        \midrule
        \multirow{3}{*}{CrossNet}
        & Base & 0.7885 & 105.18 & 0.8118 & 161.84 \\
        \cmidrule{2-6}
        & +p-DNN & 0.7930 & 163.18 & 0.8134 & 169.41 \\
        & +s-DNN & \textbf{0.7934} & \textbf{170.02} & \textbf{0.8138} & \textbf{186.59} \\
        
        \bottomrule
    \end{tabular}
    \label{tab:dnn}
\end{table}

\subsection{RQ2: Decomposition of components in DNN}\label{sec: ablation_parallel}
After establishing the role of the full DNN in mitigating dimensional collapse, we conduct a component-wise ablation to disentangle the contributions of its internal structures. 
Specifically, the DNN can be decomposed into two fundamental elements: (1) a linear-MLP module composed of stacked linear projections, and (2) non-linear activation functions that enhance representational capacity. 

To better understand their individual effects, we progressively augment the feature interaction backbone FM by first introducing the linear module and subsequently incorporating non-linear activations, thereby forming the complete p-DNN.

\paragraph{\textbf{Formulation}}
We begin with the linear module, which aggregates first-order information across feature fields through linear projections. The resulting representation is defined as
\begin{equation}
     \Phi_{\text{Linear p-DNN}} = \Phi_{FM} + 
     \mathbf{w}_{out} \cdot \mathbf{W}_{in} \cdot \text{concat}\{\mathbf{v}_j \}_{j \in \mathcal{F}_x},
\end{equation}
where $\Phi_{FM}$ denotes the output of the interaction backbone \textbf{FM}, and the second term corresponds to the linear transformation applied to the concatenated field embeddings. 
Here, $\text{concat}\{\mathbf{v}_j\}_{j \in \mathcal{F}_x}$ represents the concatenation of embeddings from all feature fields, while $\mathbf{W}_{in}$ and $\mathbf{w}_{out}$ denote the weight matrices of the linear module.

Building upon this formulation, we introduce non-linear activations to obtain the full parallel DNN:
\begin{equation}
     \Phi_{\text{p-DNN}} = \Phi_{FM} + 
     \mathbf{w}_{out} \cdot f(\mathbf{W}_{in} \cdot \text{concat}\{\mathbf{v}_j \}_{j \in \mathcal{F}_x}),
\end{equation}
where $f(\cdot)$ denotes the non-linear activation function.

\begin{figure}[t]
    \centering
    \includegraphics[width = 1.0\linewidth]{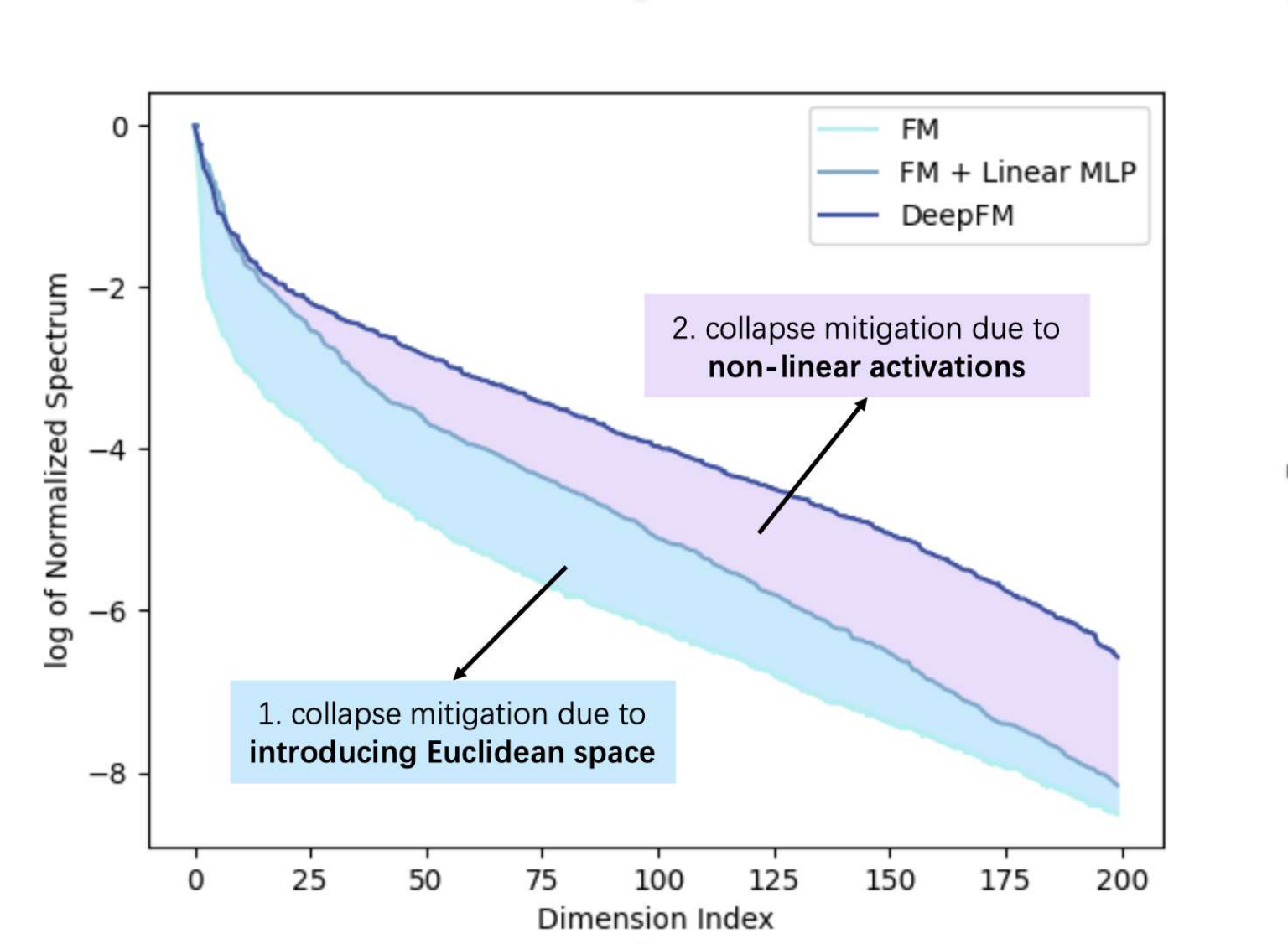}
    % \label{fig:spectrum_deepfm_truncate}
    
\caption{Singular spectrum of ablation variants on feature interaction model FM with parallel DNN, we further truncate the spectrum of FM from index 0 to 200 for clarity. Both introducing parallel linear DNNs and further employing non-linear activations can boost the singular spectrum and hence mitigate dimensional collapse.}
\label{fig:truncate_spectrum}
% \label{fig:from_FM_to_DeepFM}
\end{figure}

\paragraph{\textbf{Spectral Analysis of DNN Components}}
As illustrated in Fig.~\ref{fig:truncate_spectrum}, introducing the linear DNN leads to a noticeable elevation in the singular spectrum. 
Specifically, the top 150 singular values increase substantially compared to FM, with particularly pronounced gains among the leading 25 components. 
This spectral lifting suggests that linear projections alone expand the effective dimensionality of the embedding space and partially alleviate dimensional collapse.

Building upon the linear module, non-linear activations further reshape the spectrum by consistently elevating the singular values relative to the linear DNN.
This improvement is most pronounced between indices 25 and 200, effectively enhancing the representational capacity of the intermediate-to-large singular components.
Together, these observations lead to the following result:

\begin{tcolorbox}[colback=blue!2!white,leftrule=2.5mm,size=title] 
    \emph{\textbf{Result 2}. The progressive introduction of DNN components reveals a stepwise reduction in dimensional collapse: linear projections expand the effective embedding dimensionality, while non-linear activations further flatten the spectrum, yielding more robust representations.}
\end{tcolorbox}

\section{Probing the Effect of DNNs: A Gradient-based Analysis}

In this section, we present a gradient-based analysis to understand how parallel and stacked DNNs alleviate dimensional collapse, focusing on the representative pure feature interaction model: FM~\cite{FM2010}.
Consider a sample with categorical features $\mathcal{F}_x$, where each feature in field $i$ is mapped to an embedding vector $\mathbf{v}_i \in \mathbb{R}^k$. The model outputs a prediction $p = \sigma(\Phi)$ via logit $\Phi$ and sigmoid function, and is trained using the BCE loss $\mathcal{L}$.

For a pure FM, the output logit $\Phi_{\text{FM}} = \sum_{i<j \in \mathcal{I}_x} \langle \mathbf{v}_i \cdot \mathbf{v}_j \rangle$ explicitly captures second-order feature interactions.
The gradient of $\Phi_{\text{FM}}$ with respect to feature embedding $\mathbf{v}_i$ of field $i$ is derived as:
\begin{equation}
\frac{\partial \Phi_{\text{FM}}}{\partial \mathbf{v}_i}=\sum_{j \in \mathcal{F}_x, j \neq i} \mathbf{v}_j.
\label{eq:fm_grad}
\end{equation}

This gradient exposes a critical limitation: the update direction of $\mathbf{v}_i$ is strictly restricted to the subspace spanned by the embeddings of its interacting features $\{\mathbf{v}_j\}_{j \neq i}$. 
Such constrained gradient dynamics are consistent with the Interaction-Collapse Theory~\cite{guo2023multi-embedding}, which suggests that embeddings become bounded by low-rank fields.
This dependency progressively draws the representation space toward a lower-rank subspace, manifesting as the dimensional collapse observed in our spectral analysis.

\subsection{Analysis of Integrating DNNs}
To understand why DNNs alleviate collapse, we examine the gradient flow when introducing parallel DNNs and stacked DNNs.

\paragraph{\textbf{Parallel DNN}}
In a p-DNN architecture (e.g., DeepFM), the gradient with respect to embedding $\mathbf{v}_i$ is the sum of interaction and DNN components:
\begin{equation}
\frac{\partial \Phi_{p-DNN}}{\partial \mathbf{v}_i}=(\underbrace{\sum_{j \in \mathcal{F}_x, j \neq i} \mathbf{v}_j}_{\text {gradients \ from \  FM}}+\underbrace{\mathbf{g}_{DNN}\left(\mathbf{v}_i ; \mathcal{F}_x\right)}_{\text {gradients \ from \ DNN}}).
\label{eq:deepfm_grad}
\end{equation}
Specifically, 
\begin{equation}
\mathbf{g}_{DNN}\left(\mathbf{v}_i ; \mathcal{F}_x\right) =  (\mathbf{w}_{\text{out}} \cdot D \cdot\mathbf{W_{in}} )\cdot P_i ,
\label{eq:dnn_grad}
\end{equation}
\begin{equation}
D = \text{diag}\left(\mathbb{I}\left(\mathbf{W} \cdot \text{concat}\{\mathbf{v}_j \}_{j \in \mathcal{F}_x} + \mathbf{b} > 0\right)\right).
\end{equation}
Here, $P_i$ is a selection matrix used to extract $\mathbf{v}i$ from the concatenated input vector. $\mathbf{W}_{in}$ and $\mathbf{w}_{out}$ denote the weight matrices of the hidden and output layers, respectively, and $\mathbf{b}$ represents bias. The symbol $\mathbb{I}$ denotes the indicator function.

Besides the gradient of FM, $\mathbf{g}_{DNN}$ in Eq.~\ref{eq:deepfm_grad} introduces a highly non-linear term that fundamentally differs from the FM gradient.
This introduces orthogonal gradient components that are not present in pure interaction models, allowing embeddings to escape low-dimensional manifolds by updating in a broader vector space.

\paragraph{\textbf{Stacked DNN}}
In s-DNN architectures (e.g., NFM), the DNN acts as a non-linear filter on the interaction results, and the gradient can be formalized as:
\begin{equation}
\frac{\partial \Phi_{s-DNN}}{\partial \mathbf{v}_i}=\mathbf{g}_{DNN}\cdot \sum_{j \in \mathcal{F}_x, j \neq i} diag(\mathbf{v}_j).
\label{eq:nfm_grad}
\end{equation}
Here, $diag(\cdot)$ denotes the operator that maps a vector to a diagonal matrix, and
\begin{equation}
\mathbf{g}_{DNN} = \mathbf{w}_{out}\cdot{diag}\left(\mathbb{I}\left(\mathbf{W}_{in} \cdot \Phi_{FM}+ \mathbf{b} > 0\right)\right) \cdot \mathbf{W}_{in}.
\end{equation}

With the introduction of s-DNN, $\mathbf{g}_{DNN}$ acts as a dimension-aware weighting vector, modulating each dimension of the gradient with a distinct, non-linear signal. 
This decoupling dependency prevents dimensions from converging toward directions dominated by the low-rank subspace, thereby preserving the diversity of the embedding dimensions and yielding a flatter singular spectrum.

% \begin{figure}[t]
% \centering
% 	\subfloat[Parallel DNN\label{fig:fm_deepfm_avazu_grad}]{%
%             \includegraphics[width = 0.24\textwidth]{figure/grad/grad_spectrum_deepfm.png}
%             }
%             % \hfill
% 	\subfloat[Stacked DNN\label{fig:fm_nfm_avazu_grad}]{
%             \includegraphics[width = 0.24\textwidth]{figure/grad/grad_spectrum_nfm.png}}
% \caption{Singular value spectrum of \textbf{Gradients} backpropagated to embedding on Avazu.}
% \label{fig:grad}
% \end{figure}
\begin{figure}[t]
\centering
	\subfloat[RankMe Evolution during training\label{fig:grad_rankme}]{%
            \includegraphics[width = 0.24\textwidth]{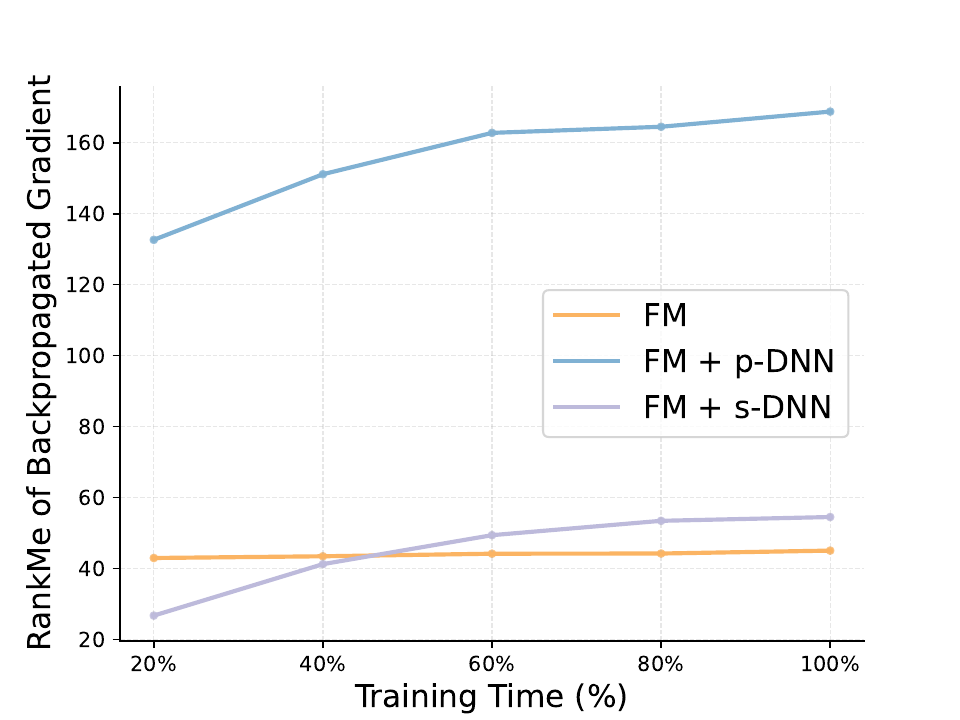}
            }
            % \hfill
	\subfloat[Singular Value Spectrum \label{fig:grad_spectrum}]{
            \includegraphics[width = 0.24\textwidth]{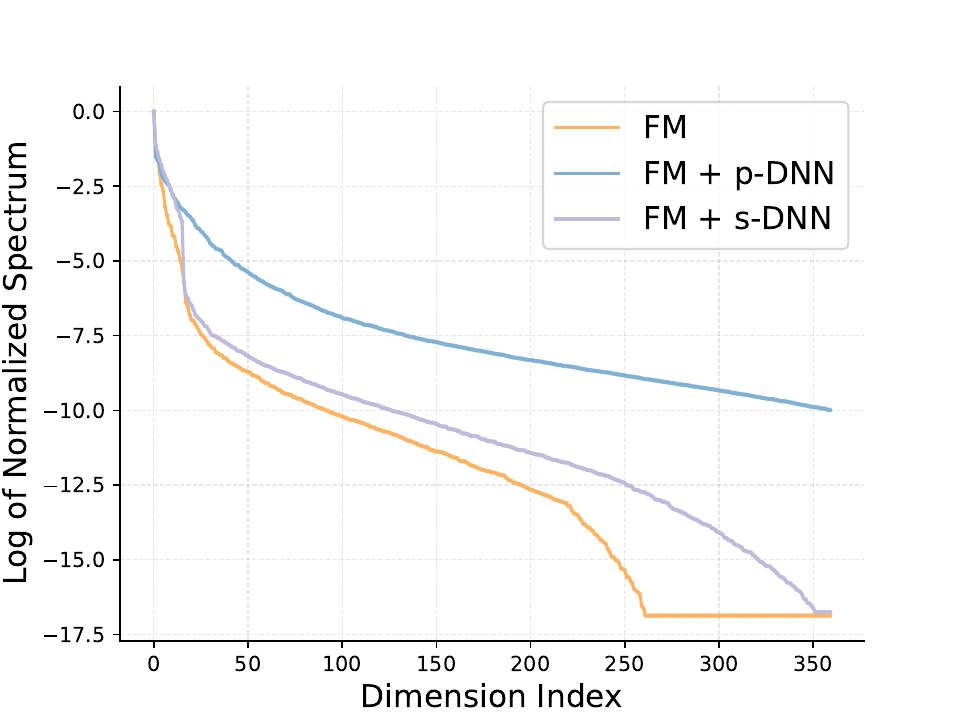}}
\caption{Spectral Analysis of \textbf{Gradients} on Avazu.}
\label{fig:grad}
\end{figure}

\subsection{Empirical Study on Gradient}
To verify our theoretical analysis of how DNNs influence gradient flow,
we conduct an empirical study on the Avazu dataset, focusing on the gradients backpropagated to feature embeddings.
We track the temporal evolution of the gradients by capturing snapshots at discrete training intervals and computing the RankMe metric to characterize the optimization trajectory in Fig.~\ref{fig:grad_rankme}.
Furthermore, we visualize the singular value spectrum of the resulting gradient matrices in Fig.~\ref{fig:grad_spectrum} to examine their effective 
dimensionality.

As illustrated in Fig.~\ref{fig:grad}, FM exhibits the lowest and nearly stationary RankMe scores, with gradient updates heavily concentrated within the top-30 singular components. 
This highly concentrated spectral distribution progressively induces dimensional collapse.
In contrast, s-DNN partially alleviates the constraints of explicit interactions via dimension-wise weighting. Most notably, p-DNN significantly boosts the RankMe values and lifts the singular values across all dimensions, enabling the model to optimize in a broader embedding space. These observations lead to the following finding:

\begin{tcolorbox}[colback=blue!2!white,leftrule=2.5mm,size=title] 
    \emph{\textbf{Findings}. DNN plays a key role in mitigating gradient-induced dimensional collapse. Empirical evidence demonstrates that DNNs alleviate dimensional collapse by fostering smoother and diverse gradient distributions during the optimization process.}
\end{tcolorbox}

\section{Conclusion}

Unlike the conventional understanding of the role of DNN in feature interaction models (\emph{i.e.} capturing implicit high-order interactions), we presented a novel embedding dimensional collapse perspective: Both parallel and stacked non-linear activated DNN mitigate the dimensional collapse of embeddings. 
% when applied to explicit feature interaction modules, 
We have also conducted comprehensive ablation studies to verify that the two components in DNN—the linear DNN component and the non-linear activation function—contribute to alleviating dimensional collapse in parallel and stacked DNN architectures.
Furthermore, we conduct an in-depth analysis from the perspective of gradients to uncover the underlying causes of dimensional collapse.

% \clearpage

\bibliographystyle{ACM-Reference-Format}
\bibliography{reference}

% \bibliographystyle{ACM-Reference-Format}
% \bibliography{reference}

\appendix

\end{document}